\documentclass{article}

\usepackage{microtype}
\usepackage{graphicx}
\usepackage{subfigure}
\usepackage{booktabs} %
\usepackage[table,dvipsnames]{xcolor}
\usepackage{epsfig}
\usepackage{pgfplotstable}
\usepackage{pgfplots}
\usepgfplotslibrary{groupplots}
\usepackage{amsmath}
\usepackage{amssymb}
\usepackage{bbm}
\usepackage{float}
\usepackage{tikz}
\usetikzlibrary{decorations.text,calc,shapes,arrows,arrows.meta, positioning,shapes.misc,decorations.markings,decorations.markings,decorations.pathreplacing,matrix}
\usepackage{booktabs}
\usepackage{multirow}
\usepackage{adjustbox}
\usepackage{verbatim}
\usepackage[T1]{fontenc}
\usepackage{adjustbox} %
\usepackage{graphicx}
\usepackage{caption}
\usepackage{paralist}%
\usepackage{siunitx}
\usepackage{nicefrac}
\usepackage{xspace}
\usepackage[colorinlistoftodos]{todonotes}

\newcommand{\textcite}[1]{\citet{#1}}

\newcommand{\lfr}{LFR\xspace}
\newcommand{\vfae}{VFA\xspace}

\usepackage[autostyle, english=american]{csquotes}
\MakeOuterQuote{"}

\usepackage[draft]{hyperref}
\usepackage[accepted]{icml2018}

\icmltitlerunning{Gradient Reversal Against Discrimination} %

\usepackage[utf8]{inputenc}

\everypar{\looseness=-1 } %
\begin{document}

\twocolumn[
\icmltitle{Gradient Reversal Against Discrimination}

\icmlsetsymbol{equal}{*}

\begin{icmlauthorlist}
\icmlauthor{Edward Raff}{bah,umbc}
\icmlauthor{Jared Sylvester}{bah}
\end{icmlauthorlist}

\icmlaffiliation{bah}{Booz Allen Hamilton}
\icmlaffiliation{umbc}{University of Maryland, Baltimore County}

\icmlcorrespondingauthor{Edward Raff}{raff\_edward@bah.com}
\icmlkeywords{Machine Learning, ICML}

\vskip 0.3in
]

\printAffiliationsAndNotice{}  %

\begin{abstract}
No methods currently exist for making arbitrary neural networks fair.
In this work we introduce GRAD, a new and simplified method to producing fair neural networks that can be used for auto-encoding fair representations or directly with predictive networks. It is easy to implement and add to existing architectures, has only one (insensitive) hyper-parameter, and provides improved individual and group fairness. We use the flexibility of GRAD to demonstrate multi-attribute protection. 
\end{abstract}

\section{Introduction}

Artificial Neural Network methods are quickly becoming ubiquitous in society, spurred by advances in image, signal, and natural language processing. This pervasiveness leads to a new need for considering the fairness of such networks from many perspectives, including: how they are used, who can access them and their training data, and potential biases in the model itself. 
There are many reasons for desiring fair classification algorithms. These include legal mandates to be non-discriminative, ensuring a moral or ethical goal, or for use as evidence in legal proceedings~\cite{Romei2014}. Despite the long-standing need and interest in this problem, there are few methods available today for training fair networks. 

When we say that a network is fair, we mean fair with respect to a protected attribute $a_p$, such as age or gender. Our desire is that a model's predicted label $\hat{y}$ given a feature vector $x$ is invariant to changes in $a_p$. An initial reaction may be to simply remove $a_p$ from the feature vector $x$. While intuitive, this "fairness through unawareness" does not remove the correlations with $a_p$ that exist in the data, and so the result will still produce a biased model~\cite{Pedreshi:2008:DDM:1401890.1401959}. 

For this reason we need to devise approaches that explicitly remove the presence of $a_p$ from the model's predictions. We do so in this work by introducing a new method to train fair neural networks.
Our approach, termed Gradient Reversal Against Discrimination (GRAD), makes use of a network which simultaneously attempts to predict the target class $y$ and protected attribute $a_p$. The key is that the gradients resulting from predictions of $a_p$ are reversed before being used for weight updates. The result is a network which is capable of learning to predict the target class but effectively inhibited from being able to predict the protected attribute.
GRAD displays competitive accuracy and improved fairness when compared to prior approaches. GRAD's advantage comes from increased simplicity compared to prior approaches, making it easier to apply and applicable to a wider class of networks.   
Prior works in this space are limited to one attribute \citep[but see][]{zafar2017fairness} and require the introduction of multiple hyper-parameters.
These parameters must be cross-validated, making the approaches challenging to use. Further, our approach can be used to augment any current model architecture, where others have been limited to auto-encoding style architectures.

\section{Gradient Reversal Against Discrimination} \label{sec:fair_nets}

We now present our new approach to developing neural networks that are fair with respect to some protected attribute. We call it \textit{Gradient Reversal Against Discrimination} (GRAD), and is inspired by recent work in transfer learning. Notably, \textcite{Ganin:2016:DTN:2946645.2946704} introduced the idea of domain adaptation by attempting to jointly predict a target label and a domain label (i.e., which domain did this data instance come from?). By treating  the protected attribute as the new domain, we can use this same approach to instead prevent the network from being biased by the protected attribute $a_p$. 

After several feature extraction layers the network forks. One branch learns to predict the target $y$, while the other attempts to predict the protected attribute $a_p$.
We term the portion of the network before the splitting point the "trunk," and those portions after the "target branch" and the "attribute branch." The final loss of the network is sum of the losses of both branches, giving $\ell(y, a_p) = \ell_t(y) + \lambda \cdot \ell_p(a_p)$.
Here, $\lambda$ determines the relative importance of fairness compared to accuracy. In practice, we find that performance is insensitive to particular choices of $\lambda$, and any value of $\lambda \in [50, 2000]$ would perform equivalently. In our experiments we will use $\lambda = 100$ without any kind of hyper-parameter optimization.

\begin{figure}[!tb]
\centering
\resizebox{0.8\columnwidth}{!}{%
\begin{tikzpicture}[auto]

      \node (x) [rectangle, draw=black,minimum width=4cm] {Raw Input \textit{x}};

      \node (fe2) [rectangle, rounded corners, draw=black, below of=x,minimum width=3.5cm] {Feature Extraction};

      \node (pred2) [rectangle, rounded corners, draw=black, below of=fe2,minimum width=3cm,xshift=-2.0cm] {Target Branch};

      \node (prot2) [rectangle, rounded corners, draw=black, below of=fe2,minimum width=3cm,xshift=+2.0cm] {Attribute Branch};
      
      \node (softPred) [rectangle, rounded corners, draw=black, below of=pred2,minimum width=3cm] {$\ell_t(y)$};
      \node (softProt) [rectangle, rounded corners, draw=black, below of=prot2,minimum width=3cm] {$\lambda \cdot \ell_p( a_p)$};

      \draw[->]
		(x) edge (fe2)  
        (pred2) edge (softPred)
        (prot2) edge (softProt);
      \draw[->, to path={-| (\tikztotarget)}]
		(fe2) edge (pred2);
    \path [draw, -latex',color=red] (fe2) -| node [midway,align=center ] {Reverse\\Gradient\\$-\frac{\partial \lambda \ell_p(a_p)}{\partial \theta_{\text{Att Branch}}}$} (prot2);

\end{tikzpicture}
}
\caption{Diagram of GRAD architecture. {\color{red} Red} connection indicates normal forward propagation, but back-propagation will reverse the signs.
}
\label{fig:fair_net}
\end{figure}
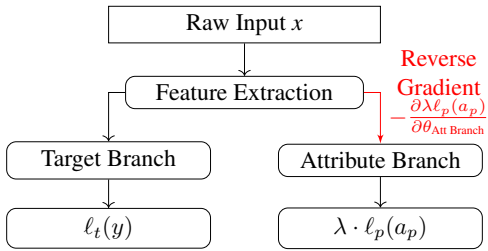

The values of both $\ell_t(y)$ and $\ell_p(a_p)$ are calculated and used to determine gradients for weight updates as usual, with one important exception. When the gradients have been back-propagated from the attribute branch they are reversed (i.e., multiplied by $-1$) before being applied to the trunk. This moves the trunk's parameters \textit{away} from optima in predictions of $a_p$, crippling the ability to correctly output the protected attribute. Since the target branch also depends on the trunk parameters, it inherits this inability to accurately output the value of the protected attribute.  No such  reversal is applied to the gradients derived from $y$, so the network's internal state representations are suitable for predicting $y$ but nescient of~$a_p$.

It is instructive to consider why it may be insufficient to set up a loss function which directly punishes the network for correctly predicting $a_p$. If this were the case, the network could achieve low loss by forming internal representation which are very good at predicting the protected attribute, and then "throw the game" by simply reversing the correct prediction in the penultimate layer. 
(That is, a potential, reliable strategy to getting the wrong answer is to become very good at getting the right answer, and then lying about what one thinks the answer should be.)
If this strategy is adopted then the representations necessary for correctly recovering $a_p$ from $x$ would be available to the target branch when making its prediction of $y$, which is the situation we aim to prevent.

\subsection*{Architecture Variants}

As mentioned above, many of the other neural approaches to fair classification take an autoencoder or representation learning approach. This approach has its advantages. For instance, it allows the person constructing the fair model to be agnostic about the ultimate task that it will be applied to. Others like ALFR consider a target value directly, and so can not be re-used for other tasks, but may perform better in practice on the specific problem they were constructed for. 

Our GRAD approach, thanks to its comparative simplicity, can be used in both formulations. This makes it the only neural network-based approach to fairness that offers both task flexibility and specificity. 

\textbf{\textit{GRAD-Auto}} will designate our approach when using an auto-encoder as the target branch's loss. That is, if $x$ is the input feature, $\tilde{x}$ will be the feature vector derived from $x$ such that the protected attribute $a_p \notin \tilde{x}$. We then use $\ell^{\text{Auto}}_t(\cdot) = ||h_{\text{target}} - \tilde{x}||^2_2$ as the loss function for the target branch, where $h_{\text{target}}$ is the activation vector from the last layer of the target branch. This approach is in the same style as LFR and VFA, where a hidden representation invariant to $a_p$ is learned, and then Logistic Regression is used on the outputs from the trunk sub-network to perform classification.

\textbf{\textit{GRAD-Pred}} will designate our task-specific approach, where we use the labels $y_i$ directly. Here we simply use the standard logistic loss $\ell^{\text{Pred}}_t(\cdot) = \log(1 + \exp(-y \cdot h_{\text{target}}))$. In this case the target branch of the network will produce a single activation, and the target branch output itself is used as the classifier directly. 

Since we are dealing with binary protected attributes, both GRAD-Auto and GRAD-Pred will have the attribute branch of the network use $\ell_p(a_p) = \log(1 + \exp(-a_p \cdot h_{\text{attribute}}))$.

In the spirit of minimizing the effort needed by the practitioner, we do not perform any hyper-parameter search for the network architecture either. Implemented in Chainer \cite{chainer_learningsys2015} we use two fully-connected layers for every branch of the network (trunk, target, \& attribute) where all hidden layers have 40 neurons. Each layer will use batch-normalization %
followed by the the ReLU activation function%
. Training is done using the Adam optimizer for gradient decent%
.
We emphasize that the heart of GRAD is the inclusion of the attribute branch with reversed gradient; this technique is flexible enough to be used regardless of the particular choices of layer types, sizes, etc. We train each model for 50 epochs, and use a validation set to select the model from the best epoch. We define best by the model having the lowest Discrimination (see \S\ref{sec:metrics}) on the validation set, breaking ties by selecting the model with the highest accuracy. When multiple attributes are protected, we use the lowest average Discrimination.

\section{Methodology} \label{sec:method}

There is currently ongoing debate about what it means for a machine learning model to be fair.
We choose to use the same evaluation procedure laid out by \citet{pmlr-v28-zemel13}. This makes our results comparable with a larger body of work, as their approach and metrics have been widely used through the literature \citep[e.g.,][]{Landeiro:2016:RTC:3015812.3015840,Bechavod2017,Dwork2017}.
We use the same evaluation procedure and metrics: Discrimination, Consistency, Delta, and Accuracy. 

\subsection{Metrics}\label{sec:metrics}

\begin{table*}[!t]
\centering
\caption{For each dataset we show Accuracy, Delta, Discrimination, and Consistency. Best results shown in \textbf{bold}, second best in \textit{italics}.}
\label{tbl:fair_results_main}
\begin{adjustbox}{max width=\textwidth}
\begin{tabular}{@{}lcccccccccccc@{}}
\toprule
      & \multicolumn{4}{c}{German}                                            & \multicolumn{4}{c}{Adult}                                             & \multicolumn{4}{c}{Health}                                            \\ 
      \cmidrule(l){2-5} \cmidrule(l){6-9}  \cmidrule(l){10-13} 
Algorithm & Acc             & Delta           & Discr           & Cons            & Acc             & Delta           & Discr           & Cons            & Acc             & Delta           & Discr           & Cons            \\ \midrule
NN-Auto   & \textit{0.7350} & 0.5334          & 0.2016          & 0.8730          & 0.7635          & 0.7191          & 0.0444          & 0.9850          & \textbf{0.8506} & 0.7939          & 0.0567          & 0.9730          \\
GRAD-Auto & 0.6750          & 0.6296          & 0.0454          & 0.8705          & 0.7554          & 0.7452          & 0.0102          & \textit{0.9924} & 0.8491          & \textbf{0.8491} & \textbf{0.0000} & \textbf{1.0000} \\
NN-Pred   & \textbf{0.7500} & 0.3637          & 0.3863          & 0.6945          & 0.7022          & 0.6268          & 0.0754          & 0.8168          & 0.8440          & 0.7511          & 0.0929          & 0.9453          \\
GRAD-Pred & 0.6750          & \textit{0.6744} & \textbf{0.0006} & \textbf{0.9705} & 0.7543          & \textit{0.7543} & \textbf{0.0000} & \textbf{1.0000} & \textit{0.8493} & 0.8486          & \textit{0.0007} & \textit{0.9999} \\ \cmidrule(l){2-13} 
LFR       & 0.5909          & 0.5867          & \textit{0.0042} & \textit{0.9408} & 0.7023          & 0.7018          & \textit{0.0006} & 0.8108          & 0.7365          & 0.7365          & \textbf{0.0000} & \textbf{1.0000} \\
VFAE      & 0.7270          & \textbf{0.6840} & 0.0430          & ---             & \textit{0.8129} & 0.7421          & 0.0708          & ---             & 0.8490          & \textit{0.8490} & \textbf{0.0000} & ---             \\
ALFR      & ---             & ---             & ---             & ---             & \textbf{0.8251} & \textbf{0.8241} & 0.0010          & ---             & ---             & ---             & ---             & ---             \\ \bottomrule
\end{tabular}
\end{adjustbox}
\end{table*}

\setlength{\abovedisplayskip}{0.5\baselineskip}
\setlength{\belowdisplayskip}{0.5\baselineskip}

Given a dataset $\{x_1, \ldots, x_n\} \in \mathcal{D}$, %
we  define the ground true label for the $i$th datum as $y_i$ and the model's prediction as $\hat{y}_i$. Each are with respect to the binary target label $y \in \{0, 1\}$. While we define both $y_i$ and $\hat{y}_i$, we emphasize that only the predicted label $\hat{y}_i$ is used in the fairness metrics. This is because fairness is not directly related to accuracy by equality of treatment. 

Discrimination is a macro-level measure of "group" fairness, and computed by the taking the  difference between the average predicted scores for each attribute value, assuming $a_p$ is a binary attribute. 
\begin{equation} \label{eq:discrim}
\text{Discrimination} = \left| \frac{\sum_{x_i \in T_{a_p}} \hat{y}_i}{|T_{a_p}|} - \frac{\sum_{x_i \in T_{\neg a_p}} \hat{y}_i }{|  T_{\neg a_p} |} \right|
\end{equation}
The second metric is Consistency, which is a micro-level measure of "individual" fairness.  For each $x_i \in \mathcal{D}$, we compare its prediction $y_i$ with the average of its $k$ nearest neighbors and take the average of this score across $\mathcal{D}$. 
\begin{equation} \label{eq:consistency}
\text{Consistency} = 1-\frac{1}{N} \sum_{i = 1}^N \left| \hat{y}_i - \frac{1}{k} \sum_{j \in k\text{-NN}(x_i)} \hat{y}_j\right|
\end{equation}

Because Consistency and Discrimination are independent of the actual accuracy of the method used, we also consider the \textit{Delta} = Accuracy $-$ Discrimination. This gives a combined measure of an algorithm's accuracy that penalizes it for biased predictions. 

We use these metrics in the same manner and on the same datasets as laid out in \textcite{pmlr-v28-zemel13} so that we can compare our results with prior work. This includes using the same training, validation, and testing splits. When training our GRAD approaches, we perform 50 epochs of training, and select the model to use from the validation performance. Specifically, we choose the epoch that had the lowest discrimination and broke ties by selecting the highest accuracy.

\subsection{Models Evaluated} \label{sec:models}

As a baseline for comparison against GRAD-Pred and GRAD-Auto, we will consider the same architecture but with the attribute branch removed. This produces a standard neural network, and will be denoted as \textit{NN}. 
For comparison with other fairness-seeking neural network algorithms, we  present prior results for 
Learning Fair Representations (\lfr) \cite{pmlr-v28-zemel13}, Variation Fair Autoencoders (\vfae) \cite{Louizos2016}, and Adversarial Learned Fair Representations (ALFR) \cite{Edwards2016} approaches.
For all models on all datasets, we report the metrics as presented in their original publications, as we were unable to replicate \vfae and ALFR's results.

\section{Results} \label{sec:results}

The results %
are given in \autoref{tbl:fair_results_main}.
For values unreported in their original work, we show a dash ("---") in the table. Our GRAD approach is shown in the top rows.
The bottom three rows include the other approaches as explained in ~\autoref{sec:models}.

When we compare the standard neural network (NN) with its GRAD counterpart, we can see that the GRAD approach \textit{always} increases the Delta and Consistency scores, and reduces the Discrimination. This shows its applicability across network types (classifying and auto-encoding). We can even see the GRAD approach improve accuracy on the Adult dataset by 5 percentage points. While we would not expect this behavior (i.e. a negative cost of fairness) in the general case, it is nonetheless interesting and it may indicate the protected attribute allows overfitting.%

Comparing the GRAD algorithms to the other neural networks LFR, VFA and ALFR, we see that GRAD is usually best or 2nd best in each metric. On both the German and Adult datasets, it achieves the best Discrimination and Consistency scores compared to any of the algorithms tested. On the German dataset VFA obtains a higher Delta score by having a high accuracy, though VFA has 4\% discrimination compared to GRAD-Pred's 0.06\%. On the Health dataset, GRAD-Auto and GRAD-Pred have near identical results.%
This is overall significantly better than the LFR approach which has an 11 percentage point difference in Accuracy and Delta scores compared to the GRAD approaches. The VFA algorithm is similarly within a fractional distance, though Consistency is not reported for VFA. 

GRAD consistently produces the highest Consistency. On the Adult dataset where VFAE and ALFR get better accuracy, it may have come at a cost of lower Consistency. This couldn't be confirmed since we could not replicate their results. 

\subsection{Multiple Protected Attributes}

In almost all prior works that we are aware, it is always assumed that there is only \textit{one} attribute that needs to be protected.  However, this is a myopic view of the world. All of the protected attributes that have been tested individually in this work,  like age, race and gender, may all co-occur and interact with each other. We show this in \autoref{tbl:diabetes_two_prot}  using the Diabetes dataset used in \textcite{Edwards2016}, which has both Race and Gender as features in the corpus. In this case GRAD-Pred and GRAD-Auto are protecting Race and Gender attributes. GRAD-Pred-R shows the results for protecting only Race, and GRAD-Pred-G shows for only protecting Gender. GRAD-Auto follows the same convention. 

\begin{table}[!tbh]
\centering
\caption{Accuracy, Delta, Discrimination (with respect to Race and Gender), and Consistency for our new method on the Diabetes dataset. 
Last four rows show GRAD models when only Race (R) or Gender (G) are protected.}
\label{tbl:diabetes_two_prot}
\begin{adjustbox}{max width=\columnwidth}
\begin{tabular}{@{}lccccc@{}}
\toprule
            &                 &                 & \multicolumn{2}{c}{Discrimination} &                 \\ \cmidrule(lr){4-5}
Algorithms  & Acc             & Delta           & Race             & Gender          & Cons            \\ \midrule
NN-Auto     & 0.5735          & 0.5392          & 0.0412           & 0.0275          & 0.6411          \\
GRAD-Auto   & 0.5765          & 0.5723          & \textit{0.0055}  & \textbf{0.0030} & 0.6288          \\
NN-Pred     & \textbf{0.6286} & \textit{0.5848} & 0.0418           & 0.0458          & \textit{0.6464} \\
GRAD-Pred   & \textit{0.5980} & \textbf{0.5949} & \textbf{0.0028}  & \textit{0.0034} & \textbf{0.7180} \\ \cmidrule(l){2-6} 
GRAD-Auto-R & 0.5851          & 0.5749          & 0.0003           & 0.0201          & 0.6404          \\
GRAD-Auto-G & 0.5640          & 0.5143          & 0.0981           & 0.0013          & 0.6093          \\
GRAD-Pred-R & 0.5844          & 0.5478          & 0.0020           & 0.0713          & 0.7538          \\
GRAD-Pred-G & 0.5941          & 0.5526          & 0.0785           & 0.0045          & 0.6849          \\ \bottomrule
\end{tabular}
\end{adjustbox}
\end{table}

Since Discrimination is computed with respect to specific attributes, in the table we show the discrimination scores with respect to both of the protected attributes. Since we have two protected attributes $a_{p^1}$ and $a_{p^2}$, we compute Delta = Accuracy $-($ Discrimination($a_{p^1}$) + Discrimination($a_{p^2}$) $)/2$. In doing so, we can see that when two protected variables are present, the GRAD approach is able to reduce Discrimination and increase Delta for both the Autoencoder and the standard softmax predictive network. GRAD-Pred also continues to increase the Consistency with respect to the naive neural network.

Comparing GRAD-Pred with GRAD-Pred-R and GRAD-Pred-G is also critical to show that protecting both attributes simultaneously provides a significant benefit. On the Diabetes data, we see the model increase its discrimination with respect to Gender when only Race is protected. Similarly, when we protect Gender, discrimination with respect to  Race increases. Explicitly protecting both is the only safe way to reduce discrimination on both.

The model shifting to leverage other protected features is not surprising. When we penalize a feature which provides information, the model must attempt to recover discriminative information in other (potentially non-linear) forms from the other features. Thus the importance and utility of GRAD to protect both simultaneously is established.

\subsection{Robustness to $\lambda$}

\begin{figure}[tb]
\centering
\begin{adjustbox}{max size={0.9\columnwidth}{0.85\textheight}}
  \begin{tikzpicture}
  \begin{groupplot}[
      group style={
        	group name=myplot, group size=2 by 1,
			vertical sep=1.5cm,%
        },
        enlarge x limits=true,
      ]
    \centering

    \nextgroupplot[title=Adult Income, %
                xmode = log,
				xlabel={$\lambda$},
				every axis plot/.append style={ultra thick}
    	]

			\addplot +[mark=none, color=red] table[x=lambda,y=GRAD_pred_adult_acc,col sep=comma]{results/lambda_robust.csv};
			\addplot +[dashed,mark=none, color=Magenta] table[x=lambda,y=GRAD_pred_adult_discrim,col sep=comma]{results/lambda_robust.csv};
            \addplot +[dash dot,mark=none, color=Turquoise] table[x=lambda,y=GRAD_pred_adult_cons,col sep=comma]{results/lambda_robust.csv};

            \addplot[dotted, mark=none, black, samples=2] coordinates {(100.0,0.0) (100.0,1.0)};
            
    \nextgroupplot[title=Heritage Health, %
                xmode = log,
				xlabel={$\lambda$},
                legend style={at={(0.03,0.5)},anchor=west},
                every axis plot/.append style={ultra thick}
    	]

			\addplot +[mark=none, color=red] table[x=lambda,y=GRAD_pred_hh_acc,col sep=comma]{results/lambda_robust.csv};
            \addlegendentry{Accuracy}
			\addplot +[dashed,mark=none, color=Magenta] table[x=lambda,y=GRAD_pred_hh_discrim,col sep=comma]{results/lambda_robust.csv};
            \addlegendentry{Discrimination}
            \addplot +[dash dot,mark=none, color=Turquoise] table[x=lambda,y=GRAD_pred_hh_cons,col sep=comma]{results/lambda_robust.csv};
            \addlegendentry{Consistency}

            \addplot[dotted, mark=none, black, samples=2] coordinates {(100.0,0.0) (100.0,1.0)};

    \end{groupplot}

  \end{tikzpicture}
\end{adjustbox}
  \caption{Plots show the performance of GRAD-Pred as a function of $\lambda$ on the x-axis (log scale).
  A dashed vertical black line shows the value $\lambda=100$ used in all experiments.
}
  \label{fig:lambda_robust}
\end{figure}
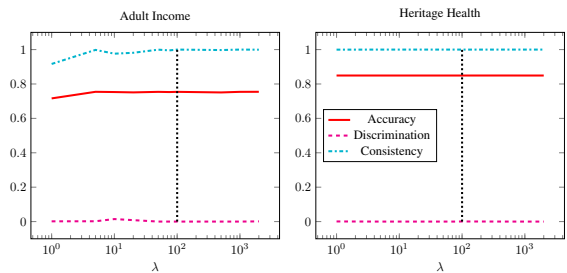

We have discussed so far that a benefit of the GRAD approach is a simplicity in application due to the having only one hyper-parameter $\lambda$. We now show that this value $\lambda$ is largely robust to the value used. In \autoref{fig:lambda_robust} we plot the Accuracy, Discrimination, and Consistency as a function of $\lambda$ for values in the range $[1, 2000]$, 
which shows GRAD's consistent behavior. On the Adult dataset, we see results stabilize after $\lambda \geq 10$. The Health dataset looks flat through the entire plot since the variation is on the order of $10^{-3}$, making it indiscernible. 
Only the Adult and Health plots are shown due to space limitations. The Diabetes plot is similar, and the German dataset has more variability due to its small size ($n=1000$). 

\section{Conclusions} \label{sec:conclusion}

We have introduced GRAD, a flexible approach for building fair neural networks that can be used to augment any general neural network, and does not mandate the auto-encoding approach of prior work or the use of cumbersome additional hyper-parameters. GRAD is competitive with prior work, can protect multiple attributes, and often delivers superior fairness through low discrimination. 

\section*{Acknowledgments}

We would like to thanks Steven Mills and Paul Terwilliger for their support of this work. 

\bibliographystyle{icml2018}
\bibliography{Mendeley,refs}

\end{document}